\documentclass{article}
\PassOptionsToPackage{authoryear, round}{natbib}
\usepackage[preprint]{neurips_2025}
\usepackage[utf8]{inputenc}
\usepackage[T1]{fontenc}

\usepackage[hidelinks]{hyperref}
\usepackage{xcolor}
\usepackage{multirow}
\usepackage{subcaption}
\usepackage{graphicx}
\usepackage{amsmath}
\usepackage{amsfonts}
\usepackage{array}
\usepackage{wrapfig}
\usepackage{cleveref}
\usepackage{microtype}
\usepackage{booktabs}
\usepackage{fontawesome5}
\usepackage{enumitem}

\usepackage{listings}
\usepackage{xcolor}
\usepackage{mathtools}

\definecolor{codegreen}{rgb}{0,0.6,0}
\definecolor{codegray}{rgb}{0.5,0.5,0.5}
\definecolor{codepurple}{rgb}{0.58,0,0.82}
\definecolor{backcolour}{rgb}{0.95,0.95,0.92}

\lstdefinestyle{mypy}{
    backgroundcolor=\color{backcolour},
    commentstyle=\color{codegreen},
    keywordstyle=\color{magenta},
    numberstyle=\tiny\color{codegray},
    stringstyle=\color{codepurple},
    basicstyle=\ttfamily\scriptsize,
    breakatwhitespace=false,
    breaklines=true,
    captionpos=b,
    keepspaces=false,
    numbersep=5pt,
    showspaces=false,
    showstringspaces=false,
    showtabs=false,
    tabsize=4,
    language=Python
}

\title{\texttt{mini-vec2vec}: Scaling Universal Geometry Alignment with Linear Transformations}
\author{Guy Dar \\ 
\texttt{guy.dar@cs.tau.ac.il} \\
Independent Researcher
}
\begin{document}

\maketitle







\begin{abstract}
We build upon \texttt{vec2vec}, a procedure designed to align embedding spaces without parallel data. While it usually finds a near-perfect alignment, it is expensive and unstable. We present \texttt{mini-vec2vec}, a simple and efficient alternative that is highly robust and incurs a much lower computational cost. Moreover, the transformation that \texttt{mini-vec2vec} learns is \textit{linear}. The method consists of three main stages: a tentative matching of pseudo-parallel embedding vectors, transformation fitting, and iterative refinement. Our linear alternative exceeds the original instantiation of \texttt{vec2vec} by orders of magnitude in efficiency, while matching or exceeding its alignment performance. The method's stability and interpretable steps facilitate scaling and unlock new opportunities for adoption in new domains and fields.
\end{abstract}
\begin{center}
    \href{https://github.com/guy-dar/mini-vec2vec}{
        \faGithub\texttt{\ github.com/guy-dar/mini-vec2vec}
    }
\end{center}

\section{Introduction}

Text embeddings, also called sentence embeddings, represent chunks of text in a high-dimensional vector space. They capture complex linguistic relationships and enable downstream tasks, such as semantic similarity and information retrieval. Naturally, different embedding models induce incompatible embedding spaces that cannot be compared directly, due to differences in architecture or initialization.  The output of one model looks like gibberish to another.
This can be easily solved with supervised training -- the translation between embedding models can be learned from embedding pairs representing the same sentences. However, the unsupervised problem, where we don't have access to parallel embeddings of the same sentences, is much harder. 

The unsupervised alignment of \textit{word} embeddings has been studied extensively for a long time, often for the practical goal of aligning monolingual word embeddings from different languages. However, \textit{text} embeddings have been mostly ignored. With the increasing popularity of databases of text embeddings, leakage of sensitive vector data becomes a real possibility. It has been shown that embedding inversion methods can extract the texts from the vectors alone \citep{song2020information, vec2text}, but they require access to the model. Unsupervised alignment can be used to extend the attack to unknown proprietary models. The attacker just translates the embeddings to a known space and applies inversion methods on the translated embeddings -- extracting the text from the data \citep{jha2025vec2vec}.  

Recently, \citet{jha2025vec2vec} have proposed \texttt{vec2vec}, a method based on CycleGAN \citep{zhu2017unpaired} for fully unsupervised text embedding alignment. As the logic behind their method, they cite a principle sometimes called \textit{Universal Geometry} or the {\textit{Platonic Representation Hypothesis}} \citep{huh2024platonic}, which posits that well-trained representations, regardless of architectures or initialization, tend to converge toward {\textit{geometrically}} similar spaces. This means that there is a way to rotate one embedding space to make it compatible with the other, and vice versa.
While \texttt{vec2vec} is successful in finding a good alignment, the use of adversarial training is computationally intensive, requires careful hyperparameter tuning, and is subject to training instabilities like mode collapse, oscillatory behavior, or failure to converge \citep{saxena2023gan_survey, goodfellow2017nips2016}. Moreover, adversarial methods typically demand substantial computational resources, including GPU acceleration and large amounts of training data -- factors that may prohibit their utilization by most researchers and users.

\begin{figure}
    \centering
    \hspace*{-0.7cm}
    \includegraphics[trim={0 5.82cm 0 0.1cm}, clip, width=1.14\linewidth]{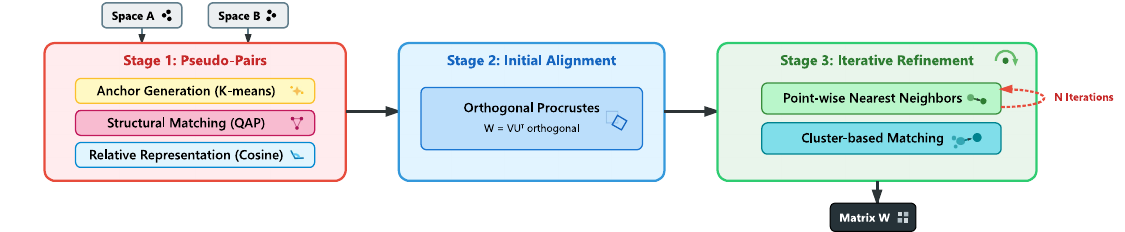}
    \caption{A simplified illustration of the \texttt{mini-vec2vec} pipeline.}
    \label{fig:pipeline}
\end{figure}

We hypothesize that the transformation between the embedding spaces is of a simple form (linear, specifically), and that it can be learned without adversarial methods. Indeed, in word embedding alignment much of the work has focused on this setup as well \citep[\textit{inter alia}]{artetxe2018vecmap,  hoshen-wolf-2018-non, grave2019unsupervised}. 
To achieve a similar goal in sentence embeddings, \texttt{mini-vec2vec} seeks to identify signals of similarities in the geometric structure of sentence embedding spaces. 
We propose a very simple pipeline (see Figure \ref{fig:pipeline} for a schematic overview), which consists of three steps: 
\begin{itemize}
    \item \textbf{Approximate Matching}: We match, in an unsupervised way, a small set of artificial pairs.
    \item \textbf{Initial Transformation}: We find a linear map between the pairs and apply it to the full data.
    \item \textbf{Iterative Refinement}: We then iteratively apply the operator, match each embedding to the closest embeddings in the other space, and learn a new operator on top of this matching.    
\end{itemize}
This pipeline resembles some methods from unsupervised word embedding alignment -- but text embeddings are somewhat different from word embeddings, so they require other tricks to make it work. Perhaps the most significant difference is the Approximate Matching method we rely on. At a high level, we need to find a new way to identify stable central landmarks and match them between sentence embedding spaces. Methods in word alignment implicitly rely on word vocabularies, which have special features (e.g., common words between languages, word frequencies, distributional properties of words, etc.). On the other hand, sentence embeddings can be seen as sparse samples from an ``infinite'' vocabulary, rendering them closer to, for instance, image embeddings.

After we have found an approximate match, the differences between word and sentence embeddings blur, and similar methods can be applied. We have developed the pipeline unaware of existing literature, so it is possible that our design decisions presented here are suboptimal or less common. Having said that, even in the Initial Alignment and Iterative Refinement steps, the design caters to specific characteristics uncommon in word embeddings, or at least less salient. For instance, when matching embeddings to their nearest neighbors, we average multiple nearest neighbors because there is no one-to-one matching between data points. This is often not the case in word alignment, since most languages, even if they are very distant, rely on closed vocabularies that significantly overlap.     

We find \texttt{mini-vec2vec} offers several key advantages: computational efficiency (very fast and runs on a commodity CPU), training stability (good performance and robust to hyperparameters), interpretability (each step has a clear geometric interpretation), and sample efficiency (60k vs 2 million samples). Through comprehensive experiments, we demonstrate that this simple approach \textbf{not only matches but often exceeds} the performance of the more complex adversarial method while requiring \textit{orders of magnitude less computational resources}.

\section{Preliminaries}
The hypothesis that neural networks converge to geometrically similar representations has been explored across multiple domains and scales. Early work by \citet{colah2015visualizing} has revealed striking structural similarities in representations of different convolutional networks, suggesting that the geometric organization of learned features might be more universal than previously assumed. The \textit{Platonic Representation Hypothesis}, formally articulated by \citet{huh2024platonic}, represents the most comprehensive theoretical framework for understanding Universal Geometry. 
The hypothesis posits that well-trained representations converge toward a shared geometric structure that reflects the underlying structure of the data domain.
This means that distances (and similarities) between objects should be roughly independent of the embedding model. For example, to compute the semantic similarity between \textit{``I love your dog''} and \textit{``I love my cat''}, we would use cosine similarity between their embeddings. According to Universal Geometry, the answer should (theoretically) have little dependence on the model used. This makes sense intuitively, as \textit{semantic} similarity originates from \textit{semantic} meaning. It is a fundamentally linguistic concept learned from the data itself. 

\paragraph{Relative Representations} \citet{moschella2022relative} have operationalized the observations of \citet{colah2015visualizing} and proposed to use {relative representations} as a mathematically principled framework for comparing latent spaces. While absolute coordinates vary arbitrarily across models, we know now that {relative information}, like similarity between two objects, is (approximately) agnostic to the choice of embedding model. They propose to choose an agreed-upon set of data points, called anchors, that other data points will be compared to, and use them like landmarks on a map. In order to specify the location of a point, we specify how close it is to every landmark. If we have enough anchors and they are sufficiently diverse, we can ``geo-locate'' the point by its relative position alone. Since it requires a small set of paired anchors, the method cannot operate as-is in a fully unsupervised manner.  

Denote the set of possible data points by $\mathcal{Z}$ (e.g., sentences, images, or any other modality; note these are abstract objects and not their vector representations) and the embedding function by $E: \mathcal{Z} \rightarrow \mathbb{R}^d$. 
Given a set of anchor data points $A = \{\mathbf{a}_i\}_{i=1}^k \subseteq \mathcal{Z}$, the \textbf{relative representation} of $\mathbf{z} \in \mathcal{Z}$ under anchors $A$ and embedding $E$ is the vector:
\begin{equation*}
\mathbf{r}_{A, E}(\mathbf{z}) = [\,\operatorname{sim}(E(\mathbf{z)}, E(\mathbf{a}_1)),\, \operatorname{sim}(E(\mathbf{z}), E(\mathbf{a}_2)),\, \ldots,\, \operatorname{sim}(E(\mathbf{z}), E(\mathbf{a}_k))\,],
\end{equation*}
where $\operatorname{sim}(\cdot, \cdot)$ is a similarity function (typically cosine similarity). 
For two embedding models satisfying the Universal Geometry assumption, relative representations of the same data point are roughly the same, i.e., $\mathbf{r}_{A, E_1} (\mathbf{z}) \approx \mathbf{r}_{A, E_2} (\mathbf{z})$. 

\section{Methodology}

\paragraph{Problem Formulation}
Following the setup laid out in \texttt{vec2vec}, let $\mathcal{X}_A = \{\mathbf{x}^A_i\}_{i=1}^{n_A}$ and $\mathcal{X}_B = \{\mathbf{x}^B_j\}_{j=1}^{n_B}$ denote two sets of $d$-dimensional embeddings. The sentences that created these embeddings are drawn from a shared underlying distribution $\mathcal{D}$, but all sentences are different and there is \textbf{{no} overlap} between the two datasets. We organize the embeddings into matrices $\mathbf{X}_A \in \mathbb{R}^{n_A \times d}$ and $\mathbf{X}_B \in \mathbb{R}^{n_B \times d}$, where each row corresponds to a single embedding vector. 
Our goal is to learn a transformation function $f: \mathbb{R}^d \to \mathbb{R}^d$ such that $f(E_A(\mathbf{z})) \approx E_B(\mathbf{z})$ where $\mathbf{z}$ represents a data point $\mathbf{z} \in \mathcal{Z}$. More generally, the setup described in \cite{jha2025vec2vec} allows for learning a pair of functions $f_A, f_B$ sending embeddings to a third space, such that $f_A(E_A(\mathbf{z})) \approx f_B(E_B(\mathbf{z}))$, but we will not take advantage of this flexibility here.

\paragraph{Motivation} 
In this work, we restrict our attention to linear transformations. Further, we constrain the search space to orthogonal transformations, i.e., matrices $\mathbf{W}$ that satisfy $\mathbf{W}\mathbf{W}^T = \mathbf{W}^T\mathbf{W} = \mathbf{I}$. This is motivated by the Universal Geometry hypothesis -- we expect distances and similarities to be preserved by the transformation. In practice, we later relax this assumption and use an exponentially-smoothed average of orthogonal matrices to enforce orthogonality softly.


\texttt{mini-vec2vec} resembles many approaches in unsupervised word embedding alignment. It proceeds in three main stages: approximate matching, transformation fitting, and iterative refinement. Unfortunately, text embeddings are somewhat more complicated than word embeddings, partly because they are {non-overlapping} -- there is no small exhaustive set (a vocabulary) of objects that need to be matched. The very concept of non-overlapping datasets is somewhat contrived for {word} embeddings, precisely because the set of possible objects is closed and relatively small. Even in multilingual matching, words mostly have a ground-truth one-to-one match. 

Beyond the overlap between the datasets, most works \citep{artetxe2016learning, artetxe2017learning, artetxe2018vecmap,hoshen-wolf-2018-non} rely on the statistics of words or other word-specific features (for finding a seed dictionary). Optimal transport-based methods \citep{alvarez2018gromov, grave2019unsupervised} rely heavily on matching points between the sets (often one-to-one, but this can be alleviated), and they become computationally intensive. Theoretically speaking, in a non-overlapping dataset, OT could find sentences that are (only) similar and match them to one another, and then a parametric mapping could be learned on top of them (a methodology similar to \citet{grave2019unsupervised}, only applied to non-overlapping data). In practice, when \citet{jha2025vec2vec} have compared their method to multiple OT methods, \textit{in an oracle setup}, they didn't work better than \textit{the most naive baseline}, indicating that even in the oracle setup, and even as a first step for learning a mapping, they are insufficient.  

We will now proceed to describe our algorithm in detail  (pseudo-code is provided in Listing~\ref{lst:full_algo}).

\begin{figure}[t]
\centering
\begin{lstlisting}[caption={Pseudo-code for \texttt{mini-vec2vec}} (refinement algorithms deferred to Listing~\ref{lst:refine}), label=lst:full_algo]
def mini_vec2vec(X_A, X_B):
    X_A, X_B = normalize(X_A), normalize(X_B)
    
    C = anchor_pairs(X_A, X_B)
    W = procrustes(C)
    
    W = refine_1(X_A, X_B, W)
    W = refine_2(X_A, X_B, W)
    
    return W

def anchor_pairs(X_A, X_B):
    R_A_list, R_B_list = [], []
    for i in range(num_runs):
        C_A = k_means(X_A, num_clusters)
        C_B = k_means(X_B, num_clusters)

        S_A = pairwise_similarities(C_A, C_A)
        S_B = pairwise_similarities(C_B, C_B)

        pi = solve_QAP(S_A, S_B)
        C_B_permuted = permute(C_B, pi)
        R_A_list.append(pairwise_similarities(X_A, C_A))
        R_B_list.append(pairwise_similarities(X_B, C_B_permuted))
    
    R_A, R_B = concatenate(R_A_list), concatenate(R_B_list)
    neighbors = knn(R_A, R_B, num_neighbors=k)
    X_matched = average_nearest_neighbors(X_B, neighbors)
    return X_A, X_matched

\end{lstlisting}
\end{figure}

\subsection{Preprocessing} 
We first normalize the embeddings in both spaces. We center each set around its respective mean and normalize the embeddings to the unit hypersphere:
\begin{align*}
\boldsymbol{\mu}_A &= \frac{1}{n_A} \mathbf{X}_A^T \mathbf{1}_{n_A}, \quad \tilde{\mathbf{X}}_A = \mathbf{X}_A - \mathbf{1}_{n_A} \boldsymbol{\mu}_A^T \\
\hat{\mathbf{X}}_A[i,:] &= \frac{\tilde{\mathbf{X}}_A[i,:]}{\|\tilde{\mathbf{X}}_A[i,:]\|_2}, \quad \text{similarly for space } B
\end{align*}
where $\boldsymbol{\mu}_A$ is the mean of space $A$, $\mathbf{1}_{n_A}$ is the vector of ones of length $n_A$, and $\|\cdot\|_2$ is the L2 norm.

\subsection{Approximate Matching}
To find an approximate matching, a small seed of pairs of (approximate) matches, we need to find stable landmarks that are likely to exist in both spaces, despite the non-overlapping data. The natural candidates are clusters. We conjecture that clusters represent recurring themes between the datasets and, therefore, should be consistent between the datasets.  But even if we find the same set of clusters in both spaces exactly, we still need to match each cluster (centroid) to its corresponding cluster (centroid) in the other space. To match the clusters, we rely on pairwise similarities, conjecturing they encode enough information to identify the matching uniquely (this intuition is shared with Gromov-Wasserstein \citep{memoli2011gromov}). We match the clusters based on similarity matrices, with Quadratic Assignment Problem (QAP) solvers. These assumptions are not trivial, but the method works relatively well as-is, though often requiring a few attempts due to the stochasticity and sensitivity of the clustering and matching algorithms. To mitigate this, we run cluster-and-match multiple times and ensemble the results. We will now describe a single run, and then explain how we robustify it by ensembling.

\paragraph{Centroid Matching} 
We perform k-means clustering in each embedding space independently and obtain cluster centroids $\{\mathbf{c}^A_j\}$ and $\{\mathbf{c}^B_j\}$ where $j = 1, \ldots, C$. We then compute pairwise similarities in each space:
\begin{align*}
S^A_{ij} &= \cos(\mathbf{c}^A_i, \mathbf{c}^A_j), \quad S^B_{ij} = \cos(\mathbf{c}^B_i, \mathbf{c}^B_j)
\end{align*}
where $\mathbf{c}^A_i$ and $\mathbf{c}^B_i$ are the $i$-th cluster centroids in spaces $A$ and $B$ respectively.

We then find the optimal matching between the cluster centroids by solving the \textit{Quadratic Assignment Problem} (QAP), which finds a permutation that aligns the similarity matrices optimally:
\begin{equation*}
\pi^*=\arg\max_{\pi\in\Pi_C}\;\sum_{i=1}^C\sum_{j=1}^C S^A_{ij}\,S^B_{\pi(i)\,\pi(j)}
\end{equation*}
where $\pi: \{1, 2, ..., C\} \rightarrow \{1, 2, ..., C\}$ is a permutation from the set $\Pi_C$ of all permutations on $C$ elements. Due to the small number of cluster centroids, this takes only a few seconds. Conceptually, we can think of QAP in terms of cosine similarity, where we want to find the permutation that will maximize the cosine similarity between the (flattened) similarity matrices.

\paragraph{Option 1: Relative Representations} Once we have identified an alignment between $A\text{-}$ and $B\text{-centroids}$, we use the aligned centroids as anchors. We represent each embedding through its relationships to anchor points. For embedding $\mathbf{x}^A_i$, we construct:
\begin{equation*}
\mathbf{r}^A_{i} = [\cos(\mathbf{x}^A_i, \mathbf{c}^A_{1}), \cos(\mathbf{x}^A_i, \mathbf{c}^A_{2}), \ldots, \cos(\mathbf{x}^A_i, \mathbf{c}^A_{C})]
\end{equation*}
Similarly, for $\mathbf{x}^B_i$, we compute (note that we need to re-arrange the indices with $\pi$):
\begin{equation*}
\mathbf{r}^B_{i} = [\cos(\mathbf{x}^B_i, \mathbf{c}^B_{\pi(1)}), \cos(\mathbf{x}^B_i, \mathbf{c}^B_{\pi(2)}), \ldots, \cos(\mathbf{x}^B_i, \mathbf{c}^B_{\pi(C)})]
\end{equation*}
\paragraph{Option 2: Alternative Approach}
Curiously, we have found that it is possible to sidestep relative representations. Instead, it is possible to use the aligned centroids directly. 
\begin{equation*}
\mathbf{r}^A_{i} = [\mathbf{c}^A_{1}, \mathbf{c}^A_{2}, \ldots, \mathbf{c}^A_{C}]; \qquad \mathbf{r}^B_{i} = [\mathbf{c}^B_{\pi(1)}, \mathbf{c}^B_{\pi(2)}, \ldots, \mathbf{c}^B_{\pi(C)}]
\end{equation*}

While the relative representations method learns from information derived from the entire data, the centroid-only approach requires just the centroids themselves. 
Both methods are efficient, and this simplified method is only slightly more efficient.  It seems to do worse after the Initial Transformation step (Section~\ref{subsec:mapping_learning}), but it converges normally in the Iterative Refinement steps (Section~\ref{subsec:refine}). We emphasize that we haven't tested it as thoroughly as the relative representation approach. For future reference, we call this variant \textbf{``\texttt{mini-vec2vec} centroid-only''}. Note that the centroid-only method is also much closer in spirit to many of the alignment methods discussed above that find a seed dictionary matching (the centroids here) and build on that. In this paper, we will \textbf{not focus} on this variant and discuss it here only for completeness.

\paragraph{Robustification via Ensembling} We improve robustness by repeating the process described above and concatenating relative representations from multiple runs:
\begin{equation*}
\mathbf{r}^A_i = [\mathbf{r}^A_{i,1}; \mathbf{r}^A_{i,2}; \ldots; \mathbf{r}^A_{i,s}] \in \mathbb{R}^{sC}, \qquad \text{similarly for space } B
\end{equation*}
This concatenation creates richer structural signatures that are more robust to the noise that comes from bad alignments.

\subsection{Initial Transformation}
\label{subsec:mapping_learning}
We construct pseudo-parallel pairs by sending each element in space $A$ to the average of its $k$ nearest neighbors from space $B$ (based on similarity in relative space):

\begin{equation*}
\mathcal{C} = \left\{ \left( \mathbf{x}^A_i, \frac{1}{k} \hspace{-0.3em} \sum_{\tiny \hspace{0.4em} {\mathbf{r}^B_j \in \mathcal{N}_k(\mathbf{r}^A_i)} \hspace{-1.8em}} \hspace{-0.3em} \mathbf{x}^B_j \,\,\,\, \right):\,  i  = 1, \ldots, n_A \right\}
\end{equation*}

where $\mathcal{N}_k(\mathbf{r}^A_i)$ denotes the $k$ nearest neighbors to $\mathbf{r}^A_i$ in the concatenated relative representation space. Note that the neighborhood information is obtained from the relative space (which is shared) and averaging takes place in the absolute space $B$. Optimal orthogonal transformation is obtained by Procrustes analysis: $\mathbf{W}^* = \mathbf{V}\mathbf{U}^T$ where $\mathbf{U}\boldsymbol{\Sigma}\mathbf{V}^T = \text{SVD}(\mathbf{A}^T\mathbf{B})$, and $\mathbf{A}$, $\mathbf{B}$ are matrices formed from the pseudo-parallel pairs in $\mathcal{C}$.

\subsection{Iterative Refinement}
\label{subsec:refine}
The initial transformation provides a coarse alignment that we refine with two complementary strategies. Pseudo-code is provided in Listing~\ref{lst:refine}.

\paragraph{Refine-1: Matching-based Refinement} This refinement strategy proceeds by transforming embeddings from space $A$ to $B$ using $\mathbf{W}$, averaging their nearest neighbors in space $B$, and obtaining a new orthogonal transformation with Procrustes analysis -- similar to the process described in the Initial Transformation stage (Section~\ref{subsec:mapping_learning}), with the crucial difference that now the similarity is computed in the ambient space $B$ rather than relative space. This technique is also present in \citet{hoshen-wolf-2018-non}, and it is often called \textit{Iterative Closest Point} \citep[ICP;][]{besl1992method}. A minor difference is that we average multiple points rather than relying on a single neighbor. As mentioned above, this is due to our no-overlap setup. Since ICP is cheap, we run it for many iterations (50--100), using subsampling to boost computational efficiency further.

We use exponential smoothing for transformation updates:
\begin{equation*}
\mathbf{W}^{(t+1)} = (1-\alpha)\mathbf{W}^{(t)} + \alpha \mathbf{W}_{\text{new}}^{(t)}
\end{equation*}

\paragraph{Refine-2: Clustering-based Refinement} This strategy aims to improve the large-scale matching between the spaces. We begin by clustering the embeddings in space $A$. We then apply $\mathbf{W}$ to the cluster centroids and cluster the $B$ embeddings, {where the clustering algorithm is initialized with the transformed $A$ centroids as the initial centroids. The transformed clusters should be close to a set of clusters in the new space. We therefore expect the clustering algorithm to make only minor adjustments, correcting biases in the transformation and moving the centroids to their ``right positions'' in space $B$. We expect the pseudo-parallel pairs based on cluster assignments to be more robust than individual nearest neighbors used in Refine-1. Similar to the first refinement strategy, we use exponential smoothing.  
Crucially, we find that \textbf{one} iteration of Refine-2 improves the alignment obtained by Refine-1. Two or more lead to a slight deterioration of the alignment. We conjecture that the ICP procedure might have introduced systematic bias that requires correction, but after this correction, the clustering-based refinement does not contribute, and only dilutes the signal.

\begin{figure}[t]
\centering
\begin{lstlisting}[caption={Pseudo-code for the iterative refinement step}, label=lst:refine]
def refine_1(X_A, X_B, W):
    for t in range(num_iters):
        X_sample = sample(X_A, subsampling_rate)
        X_transformed = X_sample @ W
        
        neighbors = knn(X_transformed, X_B, num_neighbors=k2)
        X_sample_matched = average_nearest_neighbors(X_B, neighbors)

        W_new = procrustes(X_sample, X_sample_matched)
        W = (1 - alpha) * W + alpha * W_new
    return W


def refine_2(X_A, X_B, W):
    C_A = k_means(X_A, num_clusters2)
    C_B = k_means(X_B, num_clusters2, init=C_A @ W)
    W_new = procrustes(C_A, C_B)
    
    return (1 - alpha) * W + alpha * W_new
\end{lstlisting}
\end{figure}

\section{Experiments}
We replicate the main experiment from \cite{jha2025vec2vec}. We run our algorithm for all pairs of encoders used in \cite{jha2025vec2vec}. Details about the encoders are provided in Table~\ref{tab:embedding_models} in the Appendix. Experiments use k-means clustering from scikit-learn \citep{scikit-learn} and QAP is solved with the scipy \citep{2020SciPy-NMeth} implementation of 2-OPT. {We re-run the 2-OPT alignment algorithm 30 times and choose the permutation with the highest alignment score. This helps improve the robustness of the alignment, as a single 2-OPT is often not enough.}

We take a random subset of the Natural Questions dataset \citep{natural_questions} of size $60, 000$, compute sentence embeddings with all encoders, and leave $8192$ sentences for evaluation. We split the remaining embeddings equally between the source encoder (space $A$) and target encoder (space $B$), so there is no overlap between the source and target sentences. All experiments are run on a Colab notebook with a CPU runtime and repeated three times for each pair to demonstrate the method's stability. We compare our results to the numbers reported in \citet{jha2025vec2vec}.

\paragraph{Metrics}
We evaluate alignment quality using two primary metrics. The first is \textbf{Top-1 Accuracy}, the fraction of queries where the nearest neighbor in the target space is the correct match, defined by:
\begin{equation*}
    \text{Accuracy} = \frac{1}{N} \sum_{i=1}^{N} \delta(\text{NN}(\mathbf{W} \mathbf{a}_i) = \mathbf{b}_i),
\end{equation*}
where $\delta$ is the indicator function, $\text{NN}(\cdot)$ is the nearest neighbor of source embedding $\mathbf{a}_i$ in the target space, and $\mathbf{b}_i$ is its true corresponding target embedding. The second is the \textbf{Average Rank}, the mean position of the true match in the sorted list of distances from the query to all targets,
\begin{equation*}
    \text{AvgRank} = \frac{1}{N} \sum_{i=1}^{N} \text{rank}(\mathbf{b}_i | \mathbf{a}_i).
\end{equation*}
We also report the mean \textbf{cosine similarity} after each step of the algorithm to show its progression over the stages. {Cosine similarity in itself can be used as an evaluation metric, but it can also be misleading. Cosine similarity in \texttt{mini-vec2vec} underestimates the similarity in the original space, due to the removal of the mean vector, which is a dominant part of all vectors, accounting for $\sim 30\%-70\%$ of the vector's norm. Also, setups that move the vectors to a third space, like \texttt{vec2vec}, have an advantage as they can change the geometry.} Our setup does not change the geometry (apart from aparametric translation and scaling in the preprocessing step).

\paragraph{Hyperparameters} We have found that the method is extremely robust to hyperparameter choices, as long as they are in a reasonable range. For the paper, we made an economical choice of hyperparameter values to shave time for the large number of runs, but we found the choice to be sufficiently robust even still. We use $s=30$ runs of relative representation alignment with $c=20$ clusters. In the Initial Transformation step, we use $k=50$ neighbors. We set the sampling rate in Refine-1 to $n_s = 10,000$ and utilize $k'=50$ neighbors; In Refine-2, we use $c'=500$ clusters. For exponential smoothing, we use $\alpha = 0.5$. 
The full hyperparameter configuration used in the experiments is also documented in the code linked. For the $\textit{gtr} \leftrightarrow \textit{gte}$ pair, we increase the number of clusters to $c=30$. We find that for this pair, the algorithm often converges to a sub-optimal solution in the $c=20$ regime. A detailed discussion of observations about hyperparameter choice can be found in Appendix~\ref{app:hyperparam_choice}.
\begin{table*}[t]
\centering
\scriptsize \begin{tabular}{ll|ccccccc} 
\toprule
& & \multicolumn{2}{c}{vec2vec} & \multicolumn{5}{c}{mini-vec2vec} \\
\cmidrule(lr){3-4} \cmidrule(lr){5-9}
$M_1$ & $M_2$ & Top-1 $\uparrow$ & Rank $\downarrow$ & Initial & Refine-1 & Refine-2 & Top-1 $\uparrow$ & Rank $\downarrow$ \\
\midrule
& gtr & \underline{\textbf{0.99}} & 1.19 {\scriptsize (0.1)} & 0.25 {\scriptsize (0.02)} & 0.58 {\scriptsize (0.00)} & 0.57 {\scriptsize (0.00)}& \underline{\textbf{0.99}} {\scriptsize (0.00)}  & \textbf{1.02} {\scriptsize (0.00)}  \\
gran. & stel. & \underline{0.98} & 1.05 {\scriptsize (0.0)} & 0.38 {\scriptsize (0.00)} & 0.59 {\scriptsize (0.00)} & 0.60 {\scriptsize (0.00)} & \underline{\textbf{0.99}} {\scriptsize (0.00)} & \textbf{1.03} {\scriptsize (0.00)} \\
& e5 & \underline{0.98} & 1.11 {\scriptsize (0.0)} & 0.28 {\scriptsize (0.00)} & 0.52 {\scriptsize (0.00)} & 0.53 {\scriptsize (0.00)} & \textbf{\underline{0.99}}  {\scriptsize (0.00)} & \textbf{1.05} {\scriptsize (0.00)} \\
& gte & {0.95} & 1.18 {\scriptsize (0.0)} & 0.35 {\scriptsize (0.04)} & 0.54 {\scriptsize (0.00)} & 0.56 {\scriptsize (0.00)} & \textbf{{0.98}}  {\scriptsize (0.00)} & \textbf{1.05} {\scriptsize (0.00)} \\
\midrule
& gran. & \underline{0.99} & \underline{1.02} {\scriptsize (0.0)} & 0.31 {\scriptsize (0.02)} & 0.58 {\scriptsize (0.00)} & 0.57 {\scriptsize (0.00)} & \underline{\textbf{1.00}}  {\scriptsize (0.00)} & \textbf{\underline{1.01}} {\scriptsize (0.00)} \\
gtr & stel. & \textbf{\underline{0.99}} & \textbf{\underline{1.03}} {\scriptsize (0.0)} & 0.29 {\scriptsize (0.00)} & 0.53 {\scriptsize (0.00)} & 0.54 {\scriptsize (0.00)} & {\underline{0.98}} {\scriptsize (0.00)} & \textbf{\underline{1.03}} {\scriptsize (0.00)} \\
& e5 & 0.84 & 2.88 {\scriptsize (0.2)} & 0.22 {\scriptsize (0.04)} & 0.48 {\scriptsize (0.00)} & 0.49 {\scriptsize (0.00)} & \textbf{0.98} {\scriptsize (0.00)} & \textbf{1.06} {\scriptsize (0.00)} \\
& $\text{gte}^{*}$ & {0.93} & {2.31} {\scriptsize (0.1)} & 0.29 {\scriptsize (0.02)} & 0.50 {\scriptsize (0.00)} & 0.51 {\scriptsize (0.00)} & \textbf{{0.98}}  {\scriptsize (0.00)} & \textbf{{1.05}} {\scriptsize (0.00)} \\
\midrule
& gran. & {0.95} & {1.22} {\scriptsize (0.0)} & 0.31 {\scriptsize (0.01)} & 0.57 {\scriptsize (0.00)} & 0.58 {\scriptsize (0.00)} & {\textbf{0.99}}  {\scriptsize (0.00)} & \textbf{1.03} {\scriptsize (0.00)} \\
gte & $\text{gtr}^*$ & {0.91} & {2.64} {\scriptsize (0.1)} & 0.26 {\scriptsize (0.02)} & 0.51 {\scriptsize (0.00)} & 0.52 {\scriptsize (0.00)} & \textbf{{0.98}} {\scriptsize (0.00)} & \textbf{1.04} {\scriptsize (0.00)} \\
& stella & \textbf{\underline{1.00}} & \textbf{\underline{1.00}} {\scriptsize (0.0)} & 0.42 {\scriptsize (0.02)} & 0.70 {\scriptsize (0.00)} & 0.71 {\scriptsize (0.00)} & \textbf{\underline{1.00}} {\scriptsize (0.00)} & \textbf{\underline{1.00}} {\scriptsize (0.00)} \\
& e5 & 0.99 & 5.91 {\scriptsize (0.5)} & 0.33 {\scriptsize (0.05)} & 0.60 {\scriptsize (0.00)} & 0.61 {\scriptsize (0.00)} & \textbf{1.00}  {\scriptsize (0.00)} & \textbf{1.00} {\scriptsize (0.00)} \\
\midrule
& gran. & \underline{0.98} & 1.08 {\scriptsize (0.0)} & 0.37 {\scriptsize (0.02)} & 0.58 {\scriptsize (0.00)} & 0.59 {\scriptsize (0.00)} & \underline{\textbf{0.99}} {\scriptsize (0.00)} & \textbf{1.02} {\scriptsize (0.00)} \\
stel. & gtr & \textbf{1.00} & \underline{\textbf{1.10}} {\scriptsize (0.0)} & 0.28 {\scriptsize (0.02)} & 0.54 {\scriptsize (0.00)} & 0.54 {\scriptsize (0.00)} & {0.96} {\scriptsize (0.00)} & \underline{\textbf{1.10}} {\scriptsize (0.00)} \\
& e5 & \underline{\textbf{1.00}} & \underline{\textbf{1.00}} {\scriptsize (0.0)} & 0.43 {\scriptsize (0.00)} & 0.61 {\scriptsize (0.00)} & 0.62 {\scriptsize (0.00)} & \underline{\textbf{1.00}} {\scriptsize (0.00)}& \underline{\textbf{1.00}} {\scriptsize (0.00)} \\
& gte & \underline{\textbf{1.00}} & \underline{\textbf{1.00}} {\scriptsize (0.0)} & 0.42 {\scriptsize (0.00)} & 0.69 {\scriptsize (0.00)} & 0.69 {\scriptsize (0.00)} & \underline{\textbf{1.00}}  {\scriptsize (0.00)} & \underline{\textbf{1.00}} {\scriptsize (0.00)} \\

\midrule
& gran. & \underline{\textbf{0.99}} & 2.20 {\scriptsize (0.2)} & 0.30 {\scriptsize (0.00)} & 0.54 {\scriptsize (0.00)} & 0.56 {\scriptsize (0.00)} & \underline{\textbf{0.99}} {\scriptsize (0.00)} & \textbf{1.04} {\scriptsize (0.00)} \\
e5 & gtr & 0.82 & 2.56 {\scriptsize (0.0)} & 0.22 {\scriptsize (0.03)} & 0.49 {\scriptsize (0.00)} & 0.51 {\scriptsize (0.00)} & \textbf{0.96} {\scriptsize (0.00)} & \textbf{1.10} {\scriptsize (0.00)} \\
& stel. & \underline{\textbf{1.00}} & \textbf{\underline{1.00}} {\scriptsize (0.0)} & 0.44 {\scriptsize (0.01)} & 0.64 {\scriptsize (0.00)} & 0.65 {\scriptsize (0.00)} & \underline{\textbf{1.00}} {\scriptsize (0.00)} & \underline{\textbf{1.00}} {\scriptsize (0.00)} \\
& gte & \textbf{\underline{1.00}} & \textbf{\underline{1.00}} {\scriptsize (0.0)} & 0.33 {\scriptsize (0.00)} & 0.61 {\scriptsize (0.00)} & 0.62 {\scriptsize (0.00)} & \underline{\textbf{1.00}}  {\scriptsize (0.00)} & \underline{\textbf{1.00}} {\scriptsize (0.00)} \\
\bottomrule
\end{tabular}
\caption{\small Results show mean $\pm$ standard deviation over 3 runs. Bold indicates best performance. Underline indicates nearly equally good performance ($\leq 0.01$ difference). Bold and underline combined indicate better performance, but only within a $0.01$ margin.\\
{${}^*$ For the $gtr \leftrightarrow gte$ pair, we modify the number of clusters to $c=30$ in the approximate matching stage.  
}}
\label{tab:results}

\end{table*}

\paragraph{Results} From the results in Table~\ref{tab:results}, we observe that our approach matches or exceeds \texttt{vec2vec} across (almost) the entire table. The improvement is most noticeable in the rank metric. Importantly, the method is extremely robust as is evidenced by the small standard deviation, already by the end of the first refinement step. Unlike \texttt{vec2vec}, our method does not collapse on the $\textit{e5} \leftrightarrow \textit{gtr}$ pair or the $\textit{gtr} \leftrightarrow\textit{gte}$ pair. 
We underline scores that are within a $0.01$ margin, as they might result from rounding noise, stochasticity, and a slightly different experimental setup.\footnote{Our results are calculated over multiple runs and a single evaluation set, while vec2vec computes the results of a single run over 8 evaluation sets.} Due to the small variance in most experiments, in both methods, we believe that a $0.01$ margin is good enough. 

\paragraph{Analysis} The slightly superior results of \texttt{mini-vec2vec} are very encouraging, but even in pairs where the results are slightly worse or match \texttt{vec2vec}, the main point is that this is achieved for a fraction of the cost.  One run of \texttt{mini-vec2vec} is completed on a CPU in \textbf{less than ten minutes}, while \texttt{vec2vec} requires 1--7 days on a GPU, depending on the hardware. Another important point from the point of view of robustness is that even where the \texttt{vec2vec} results are relatively low, \texttt{mini-vec2vec} still performs well, close to perfect performance.
Analyzing the convergence of our algorithm, we observe that despite variation in the Initial Transformation step (Initial), Refine-1 consistently improves cosine similarity and converges to stationary points with nearly identical cosine similarity values across different initializations. This reproducible convergence behavior suggests the optimization landscape is well-structured.
The convergence of Refine-1 appears to reach a stationary point of the implicit nearest-neighbor matching objective. Refine-2 successfully improves upon this converged solution, which might indicate that the converged solution has some systematic bias.

\section{Discussion}
\label{sec:discussion}
\paragraph{Failure Modes}
We find that despite the many experiments we ran, ranging diverse hyperparameter configurations, failure modes were extremely rare. In fact, we didn't experience \textit{any} except on the $gte \leftrightarrow gtr$ pair, and modifying the number of clusters to $c=30$ solved this, though experimentation with more runs has shown that failure modes still exist in the \textit{gtr} $\rightarrow$ \textit{gte} direction, they are just rarer. Crucially, failure modes are \textit{not} catastrophic. Failure modes are characterized by worse top-1 accuracy and average rank -- but the decrease is not catastrophic, leading to $\sim90\%$ top-1 accuracy and $5-10$ average rank, much better than any naive baseline.


\paragraph{Main Takeaways} We would like to emphasize the subtle points that make the method work. While word alignment methods weren't suitable in this setup, they are likely not completely irrelevant. The main drivers of success are stable landmarks and tractable matching. We chose to focus on clusters as these landmarks. However, it is not the only option. A preliminary study \citep{yamagiwa2023discovering} on word and image embeddings has found that Independent Component Analysis (ICA)  can identify shared interpretable directions between embedding spaces. We don't preclude the possibility that they could have been used instead. The important part is to find anchors that are consistent despite zero overlap. This gives us a match between \textit{sets}, and leaves us with the task of aligning the individual anchors. The fact that the number of anchors is relatively small makes the process very fast, but also makes the problem \textit{simpler}. It is easier to find near-optimal solutions when the number of samples is not large, while the problem quickly gets out of hand due to its combinatorial nature. We used QAP for matching, which is closely related to Optimal Transport. The small scale does more than speed up the process, but likely also improves the performance of the algorithm. We believe this might have worked with Gromov-Wasserstein and other OT algorithms as well, as long as the same principles were kept. Last but not least, robustness to noise is another crucial driver of success. While \citet{hoshen-wolf-2018-non} re-initialize their pipeline (500 times, which is very expensive, though parallelizable) to fight noise and arbitrariness, we use many noisy assignments in one run, addressing the problem without multiple runs. Other tricks, not necessarily common in literature, such as exponentially-smoothed orthogonal matrices, cluster-based matching, and others, have empirically contributed to boosting performance significantly, but even without them performance was decent.

\section{Related Work}
 \paragraph{Universality} The study of universality has its roots in several works. Representation alignment is well-studied in the supervised setup.~\citet{lenc2015stitching} and \citet{bansal2021revisiting} have demonstrated that linear transformations are often sufficient for aligning neural network representations in a procedure they called \textit{stitching}. \cite{mikolov2013exploiting} have identified that word embeddings from different languages converge to similar geometries and used it to align them with a learned linear transformation from a seed dictionary. \cite{colah2015visualizing} visualized representations from different convolutional neural networks using relative representations. \cite{moschella2022relative} have used these insights to transition between representation spaces through a universal space, utilizing it for zero-shot stitching. Concurrently, \cite{dar2023analyzing} have demonstrated zero-shot stitching using (a method equivalent to) relative representations with token embedding anchors. 
 \cite{huh2024platonic} have aggregated and unified previous work under the name \textit{Platonic Representation Hypothesis}.

\paragraph{Unsupervised Alignment} The task of unsupervised alignment of embeddings, too,  has a long history in word embeddings \citep{conneau2017word, chen-cardie-2018-unsupervised, zhang-etal-2017-adversarial} and machine translation \citep{lample2017unsupervised}. These early methods relied on adversarial training. Artetxe et al. have started a line of supervised and semi-supervised alignment methods relying on an iterative refinement process \citep{artetxe2016learning, artetxe2017learning}, which culminated in VecMap \citep{artetxe2018vecmap}, an unsupervised method that relied on the distributional features of words. Another unsupervised work \citep{hoshen-wolf-2018-non} used a three-step algorithm not unlike our own, but it required hundreds of re-initializations and used word frequencies. The works of \cite{alvarez2018gromov} and \cite{grave2019unsupervised} have utilized variants of optimal transport to align word embeddings. They require the existence of a match between points, at least approximately, in order to work. Works on unsupervised alignment in other domains, for example, image or cross-modal image-text alignment, have mostly focused on semi-supervised setups and overlapping datasets \citep{schnaus2025,maniparambil2024}, which inspired our initial attempts.

\section{Conclusion}

We have presented \texttt{mini-vec2vec}, a linear approach to unsupervised embedding alignment that achieves competitive performance while offering substantial advantages in computational efficiency, training stability, and interpretability. Our method demonstrates that adversarial training may be unnecessary for reliable embedding alignment, and identifying structural correspondences is possible in other ways. The success of our approach provides another strong empirical evidence for the Universal Geometry hypothesis and suggests that the geometric regularities in learned representations are more robust and accessible than previously assumed. By decomposing the alignment problem into classical optimization components, we achieve both theoretical clarity and practical efficiency.
The efficiency and accessibility of our method can help democratize research on embedding alignment and cross-modal representation learning. By reducing computational barriers, we enable broader participation in this research area and facilitate deployment in resource-constrained environments. From a security perspective, efficient alignment methods raise important considerations about data privacy and intellectual property protection in embedding spaces. If alignment can be performed quickly and reliably, it may become easier to leak data in the scenario described in \citet{jha2025vec2vec}.

\section*{Acknowledgements}
We thank John X. Morris for useful feedback. We thank Google for making CPU and GPU notebooks freely available.

\bibliographystyle{plainnat}
\bibliography{references}

\begin{thebibliography}{36}
\providecommand{\natexlab}[1]{#1}
\providecommand{\url}[1]{\texttt{#1}}
\expandafter\ifx\csname urlstyle\endcsname\relax
  \providecommand{\doi}[1]{doi: #1}\else
  \providecommand{\doi}{doi: \begingroup \urlstyle{rm}\Url}\fi

\bibitem[Alvarez-Melis and Jaakkola(2018)]{alvarez2018gromov}
David Alvarez-Melis and Tommi Jaakkola.
\newblock Gromov-wasserstein alignment of word embedding spaces.
\newblock In \emph{Proceedings of the 2018 conference on empirical methods in natural language processing}, pages 1881--1890, 2018.

\bibitem[Artetxe et~al.(2016)Artetxe, Labaka, and Agirre]{artetxe2016learning}
Mikel Artetxe, Gorka Labaka, and Eneko Agirre.
\newblock Learning principled bilingual mappings of word embeddings while preserving monolingual invariance.
\newblock In \emph{Proceedings of the 2016 conference on empirical methods in natural language processing}, pages 2289--2294, 2016.

\bibitem[Artetxe et~al.(2017)Artetxe, Labaka, and Agirre]{artetxe2017learning}
Mikel Artetxe, Gorka Labaka, and Eneko Agirre.
\newblock Learning bilingual word embeddings with (almost) no bilingual data.
\newblock In \emph{Proceedings of the 55th Annual Meeting of the Association for Computational Linguistics (Volume 1: Long Papers)}, pages 451--462, 2017.

\bibitem[Artetxe et~al.(2018)Artetxe, Labaka, and Agirre]{artetxe2018vecmap}
Mikel Artetxe, Gorka Labaka, and Eneko Agirre.
\newblock A robust self-learning method for fully unsupervised cross-lingual mappings of word embeddings.
\newblock \emph{arXiv preprint arXiv:1805.06297}, 2018.

\bibitem[Awasthy et~al.(2025)Awasthy, Trivedi, Li, Bornea, Cox, Daniels, Franz, Goodhart, Iyer, Kumar, Lastras, McCarley, Murthy, P, Rosenthal, Roukos, Sen, Sharma, Sil, Soule, Sultan, and Florian]{granite}
Parul Awasthy, Aashka Trivedi, Yulong Li, Mihaela Bornea, David Cox, Abraham Daniels, Martin Franz, Gabe Goodhart, Bhavani Iyer, Vishwajeet Kumar, Luis Lastras, Scott McCarley, Rudra Murthy, Vignesh P, Sara Rosenthal, Salim Roukos, Jaydeep Sen, Sukriti Sharma, Avirup Sil, Kate Soule, Arafat Sultan, and Radu Florian.
\newblock Granite embedding models, 2025.
\newblock URL \url{https://arxiv.org/abs/2502.20204}.

\bibitem[Bansal et~al.(2021)Bansal, Nakkiran, and Barak]{bansal2021revisiting}
Yamini Bansal, Preetum Nakkiran, and Boaz Barak.
\newblock Revisiting model stitching to compare neural representations.
\newblock \emph{Advances in neural information processing systems}, 34:\penalty0 225--236, 2021.

\bibitem[Besl and McKay(1992)]{besl1992method}
Paul~J Besl and Neil~D McKay.
\newblock Method for registration of 3-d shapes.
\newblock In \emph{Sensor fusion IV: control paradigms and data structures}, volume 1611, pages 586--606. Spie, 1992.

\bibitem[Chen and Cardie(2018)]{chen-cardie-2018-unsupervised}
Xilun Chen and Claire Cardie.
\newblock Unsupervised multilingual word embeddings.
\newblock In Ellen Riloff, David Chiang, Julia Hockenmaier, and Jun{'}ichi Tsujii, editors, \emph{Proceedings of the 2018 Conference on Empirical Methods in Natural Language Processing}, pages 261--270, Brussels, Belgium, October-November 2018. Association for Computational Linguistics.
\newblock \doi{10.18653/v1/D18-1024}.
\newblock URL \url{https://aclanthology.org/D18-1024/}.

\bibitem[Conneau et~al.(2017)Conneau, Lample, Ranzato, Denoyer, and J{\'e}gou]{conneau2017word}
Alexis Conneau, Guillaume Lample, Marc'Aurelio Ranzato, Ludovic Denoyer, and Herv{\'e} J{\'e}gou.
\newblock Word translation without parallel data.
\newblock \emph{arXiv preprint arXiv:1710.04087}, 2017.

\bibitem[Dar et~al.(2023)Dar, Geva, Gupta, and Berant]{dar2023analyzing}
Guy Dar, Mor Geva, Ankit Gupta, and Jonathan Berant.
\newblock Analyzing transformers in embedding space.
\newblock In \emph{Proceedings of the 61st Annual Meeting of the Association for Computational Linguistics (Volume 1: Long Papers)}, pages 16124--16170, 2023.

\bibitem[Goodfellow(2017)]{goodfellow2017nips2016}
Ian Goodfellow.
\newblock Nips 2016 tutorial: Generative adversarial networks, 2017.
\newblock URL \url{https://arxiv.org/abs/1701.00160}.

\bibitem[Grave et~al.(2019)Grave, Joulin, and Berthet]{grave2019unsupervised}
Edouard Grave, Armand Joulin, and Quentin Berthet.
\newblock Unsupervised alignment of embeddings with wasserstein procrustes.
\newblock In \emph{The 22nd International Conference on Artificial Intelligence and Statistics}, pages 1880--1890. PMLR, 2019.

\bibitem[Hoshen and Wolf(2018)]{hoshen-wolf-2018-non}
Yedid Hoshen and Lior Wolf.
\newblock Non-adversarial unsupervised word translation.
\newblock In Ellen Riloff, David Chiang, Julia Hockenmaier, and Jun{'}ichi Tsujii, editors, \emph{Proceedings of the 2018 Conference on Empirical Methods in Natural Language Processing}, pages 469--478, Brussels, Belgium, October-November 2018. Association for Computational Linguistics.
\newblock \doi{10.18653/v1/D18-1043}.
\newblock URL \url{https://aclanthology.org/D18-1043/}.

\bibitem[Huh et~al.(2024)Huh, Cheung, Wang, and Isola]{huh2024platonic}
Minyoung Huh, Brian Cheung, Tongzhou Wang, and Phillip Isola.
\newblock The platonic representation hypothesis.
\newblock \emph{arXiv preprint arXiv:2405.07987}, 2024.

\bibitem[Jha et~al.(2025)Jha, Zhang, Shmatikov, and Morris]{jha2025vec2vec}
Rishi Jha, Collin Zhang, Vitaly Shmatikov, and John~X Morris.
\newblock Harnessing the universal geometry of embeddings.
\newblock \emph{arXiv preprint arXiv:2505.12540}, 2025.

\bibitem[Kwiatkowski et~al.(2019)Kwiatkowski, Palomaki, Redfield, Collins, Parikh, Alberti, Epstein, Polosukhin, Devlin, Lee, Toutanova, Jones, Kelcey, Chang, Dai, Uszkoreit, Le, and Petrov]{natural_questions}
Tom Kwiatkowski, Jennimaria Palomaki, Olivia Redfield, Michael Collins, Ankur Parikh, Chris Alberti, Danielle Epstein, Illia Polosukhin, Jacob Devlin, Kenton Lee, Kristina Toutanova, Llion Jones, Matthew Kelcey, Ming-Wei Chang, Andrew~M. Dai, Jakob Uszkoreit, Quoc Le, and Slav Petrov.
\newblock Natural questions: A benchmark for question answering research.
\newblock \emph{Transactions of the Association for Computational Linguistics}, 7:\penalty0 452--466, 2019.
\newblock \doi{10.1162/tacl_a_00276}.
\newblock URL \url{https://aclanthology.org/Q19-1026/}.

\bibitem[Lample et~al.(2017)Lample, Conneau, Denoyer, and Ranzato]{lample2017unsupervised}
Guillaume Lample, Alexis Conneau, Ludovic Denoyer, and Marc'Aurelio Ranzato.
\newblock Unsupervised machine translation using monolingual corpora only.
\newblock \emph{arXiv preprint arXiv:1711.00043}, 2017.

\bibitem[Lenc and Vedaldi(2015)]{lenc2015stitching}
Karel Lenc and Andrea Vedaldi.
\newblock Understanding image representations by measuring their equivariance and equivalence.
\newblock In \emph{Proceedings of the IEEE conference on computer vision and pattern recognition}, pages 991--999, 2015.

\bibitem[Li et~al.(2023)Li, Zhang, Zhang, Long, Xie, and Zhang]{gte}
Zehan Li, Xin Zhang, Yanzhao Zhang, Dingkun Long, Pengjun Xie, and Meishan Zhang.
\newblock Towards general text embeddings with multi-stage contrastive learning, 2023.
\newblock URL \url{https://arxiv.org/abs/2308.03281}.

\bibitem[Maniparambil et~al.(2024)Maniparambil, Akshulakov, Djilali, El~Amine~Seddik, Narayan, Mangalam, and O'Connor]{maniparambil2024}
Mayug Maniparambil, Raiymbek Akshulakov, Yasser Abdelaziz~Dahou Djilali, Mohamed El~Amine~Seddik, Sanath Narayan, Karttikeya Mangalam, and Noel~E O'Connor.
\newblock Do vision and language encoders represent the world similarly?
\newblock In \emph{Proceedings of the IEEE/CVF Conference on Computer Vision and Pattern Recognition}, pages 14334--14343, 2024.

\bibitem[M{\'e}moli(2011)]{memoli2011gromov}
Facundo M{\'e}moli.
\newblock Gromov--wasserstein distances and the metric approach to object matching.
\newblock \emph{Foundations of computational mathematics}, 11\penalty0 (4):\penalty0 417--487, 2011.

\bibitem[Mikolov et~al.(2013)Mikolov, Le, and Sutskever]{mikolov2013exploiting}
Tomas Mikolov, Quoc~V Le, and Ilya Sutskever.
\newblock Exploiting similarities among languages for machine translation.
\newblock \emph{arXiv preprint arXiv:1309.4168}, 2013.

\bibitem[Morris et~al.(2023)Morris, Kuleshov, Shmatikov, and Rush]{vec2text}
John Morris, Volodymyr Kuleshov, Vitaly Shmatikov, and Alexander~M Rush.
\newblock Text embeddings reveal (almost) as much as text.
\newblock In \emph{Proceedings of the 2023 Conference on Empirical Methods in Natural Language Processing}, pages 12448--12460, 2023.

\bibitem[Moschella et~al.(2022)Moschella, Maiorca, Fumero, Norelli, Locatello, and Rodol{\`a}]{moschella2022relative}
Luca Moschella, Valentino Maiorca, Marco Fumero, Antonio Norelli, Francesco Locatello, and Emanuele Rodol{\`a}.
\newblock Relative representations enable zero-shot latent space communication.
\newblock \emph{arXiv preprint arXiv:2209.15430}, 2022.

\bibitem[Ni et~al.(2021)Ni, Qu, Lu, Dai, Ábrego, Ma, Zhao, Luan, Hall, Chang, and Yang]{gtr}
Jianmo Ni, Chen Qu, Jing Lu, Zhuyun Dai, Gustavo~Hernández Ábrego, Ji~Ma, Vincent~Y. Zhao, Yi~Luan, Keith~B. Hall, Ming-Wei Chang, and Yinfei Yang.
\newblock Large dual encoders are generalizable retrievers, 2021.
\newblock URL \url{https://arxiv.org/abs/2112.07899}.

\bibitem[Olah(2015)]{colah2015visualizing}
Christopher Olah.
\newblock Visualizing representations: Deep learning and human beings, 2015.
\newblock URL \url{http://colah.github.io/posts/2015-01-Visualizing-Representations/}.

\bibitem[Pedregosa et~al.(2011)Pedregosa, Varoquaux, Gramfort, Michel, Thirion, Grisel, Blondel, Prettenhofer, Weiss, Dubourg, Vanderplas, Passos, Cournapeau, Brucher, Perrot, and Duchesnay]{scikit-learn}
F.~Pedregosa, G.~Varoquaux, A.~Gramfort, V.~Michel, B.~Thirion, O.~Grisel, M.~Blondel, P.~Prettenhofer, R.~Weiss, V.~Dubourg, J.~Vanderplas, A.~Passos, D.~Cournapeau, M.~Brucher, M.~Perrot, and E.~Duchesnay.
\newblock Scikit-learn: Machine learning in {P}ython.
\newblock \emph{Journal of Machine Learning Research}, 12:\penalty0 2825--2830, 2011.

\bibitem[Saxena and Cao(2023)]{saxena2023gan_survey}
Divya Saxena and Jiannong Cao.
\newblock Generative adversarial networks (gans survey): Challenges, solutions, and future directions, 2023.
\newblock URL \url{https://arxiv.org/abs/2005.00065}.

\bibitem[Schnaus et~al.(2025)Schnaus, Araslanov, and Cremers]{schnaus2025}
Dominik Schnaus, Nikita Araslanov, and Daniel Cremers.
\newblock It's a (blind) match! towards vision-language correspondence without parallel data.
\newblock In \emph{Proceedings of the Computer Vision and Pattern Recognition Conference}, pages 24983--24992, 2025.

\bibitem[Song and Raghunathan(2020)]{song2020information}
Congzheng Song and Ananth Raghunathan.
\newblock Information leakage in embedding models.
\newblock In \emph{Proceedings of the 2020 ACM SIGSAC conference on computer and communications security}, pages 377--390, 2020.

\bibitem[Virtanen et~al.(2020)Virtanen, Gommers, Oliphant, Haberland, Reddy, Cournapeau, Burovski, Peterson, Weckesser, Bright, {van der Walt}, Brett, Wilson, Millman, Mayorov, Nelson, Jones, Kern, Larson, Carey, Polat, Feng, Moore, {VanderPlas}, Laxalde, Perktold, Cimrman, Henriksen, Quintero, Harris, Archibald, Ribeiro, Pedregosa, {van Mulbregt}, and {SciPy 1.0 Contributors}]{2020SciPy-NMeth}
Pauli Virtanen, Ralf Gommers, Travis~E. Oliphant, Matt Haberland, Tyler Reddy, David Cournapeau, Evgeni Burovski, Pearu Peterson, Warren Weckesser, Jonathan Bright, St{\'e}fan~J. {van der Walt}, Matthew Brett, Joshua Wilson, K.~Jarrod Millman, Nikolay Mayorov, Andrew R.~J. Nelson, Eric Jones, Robert Kern, Eric Larson, C~J Carey, {\.I}lhan Polat, Yu~Feng, Eric~W. Moore, Jake {VanderPlas}, Denis Laxalde, Josef Perktold, Robert Cimrman, Ian Henriksen, E.~A. Quintero, Charles~R. Harris, Anne~M. Archibald, Ant{\^o}nio~H. Ribeiro, Fabian Pedregosa, Paul {van Mulbregt}, and {SciPy 1.0 Contributors}.
\newblock {{SciPy} 1.0: Fundamental Algorithms for Scientific Computing in Python}.
\newblock \emph{Nature Methods}, 17:\penalty0 261--272, 2020.
\newblock \doi{10.1038/s41592-019-0686-2}.
\newblock URL \url{https://doi.org/10.1038/s41592-019-0686-2}.

\bibitem[Wang et~al.(2024)Wang, Yang, Huang, Jiao, Yang, Jiang, Majumder, and Wei]{e5}
Liang Wang, Nan Yang, Xiaolong Huang, Binxing Jiao, Linjun Yang, Daxin Jiang, Rangan Majumder, and Furu Wei.
\newblock Text embeddings by weakly-supervised contrastive pre-training, 2024.
\newblock URL \url{https://arxiv.org/abs/2212.03533}.

\bibitem[Yamagiwa et~al.(2023)Yamagiwa, Oyama, and Shimodaira]{yamagiwa2023discovering}
Hiroaki Yamagiwa, Momose Oyama, and Hidetoshi Shimodaira.
\newblock Discovering universal geometry in embeddings with ica.
\newblock \emph{arXiv preprint arXiv:2305.13175}, 2023.

\bibitem[Zhang et~al.(2025)Zhang, Li, Zeng, and Wang]{stella}
Dun Zhang, Jiacheng Li, Ziyang Zeng, and Fulong Wang.
\newblock Jasper and stella: distillation of sota embedding models, 2025.
\newblock URL \url{https://arxiv.org/abs/2412.19048}.

\bibitem[Zhang et~al.(2017)Zhang, Liu, Luan, and Sun]{zhang-etal-2017-adversarial}
Meng Zhang, Yang Liu, Huanbo Luan, and Maosong Sun.
\newblock Adversarial training for unsupervised bilingual lexicon induction.
\newblock In Regina Barzilay and Min-Yen Kan, editors, \emph{Proceedings of the 55th Annual Meeting of the Association for Computational Linguistics (Volume 1: Long Papers)}, pages 1959--1970, Vancouver, Canada, July 2017. Association for Computational Linguistics.
\newblock \doi{10.18653/v1/P17-1179}.
\newblock URL \url{https://aclanthology.org/P17-1179/}.

\bibitem[Zhu et~al.(2017)Zhu, Park, Isola, and Efros]{zhu2017unpaired}
Jun-Yan Zhu, Taesung Park, Phillip Isola, and Alexei~A Efros.
\newblock Unpaired image-to-image translation using cycle-consistent adversarial networks.
\newblock In \emph{Proceedings of the IEEE international conference on computer vision}, pages 2223--2232, 2017.

\end{thebibliography}

\appendix
\section{Additional Assets}
\begin{table}[h]
  \centering
  \scriptsize
  \begin{tabular}{llllrr}
    \toprule
    Model & Params & Backbone & Paper & Year\\
    \midrule
    gtr     & 110M & T5     & \citeauthor{gtr} & 2021 \\
    e5      & 109M & BERT   & \citeauthor{e5} & 2022 \\
    gte     & 109M & BERT    & \citeauthor{gte} & 2023\\
    stella  & 109M & BERT   & \citeauthor{stella} & 2023 \\
    granite & 278M & RoBERTa& \citeauthor{granite} & 2024 \\
    \bottomrule
  \end{tabular}
  \caption{Embedding models used in our experiments}
  \label{tab:embedding_models}
\end{table}
\section{Choice of Hyperparameters}
\label{app:hyperparam_choice}
\texttt{mini-vec2vec} is stable under different choices of parameters, often converging to similar stationary points. Moreover, unlike many machine learning algorithms, most hyperparameters have a clear interpretation and often have a monotone behavior. Here we share trends that we have observed empirically in some setups, but a rigorous experimental analysis has not been made.
\begin{itemize}
\item \textbf{Number of concatenated representations}: Increasing the number of runs in the anchor discovery step reduces the number of the failure modes.  This is because noisy assignments are diluted, emphasizing the signal from true assignments. Note that even with very few runs, good alignments can be reached at random but with smaller probability.  
\item \textbf{Number of iterations}: We find that the alignment is monotonically increasing as the number of iterations grows, unlike some algorithms where increasing information might cause degradation. This makes it a safe bet to increase the number of iterations. Across all cases examined, though, we found convergence to be quite fast. Almost without exception, 100 iterations were enough by a large margin. 
\item \textbf{Subsampling rate}: Subsampling allows us to speed up the algorithm quite substantially, but it can increase noise. The larger the sample size is, the closer this approximation is to the data. As a rule of thumb, we generally considered higher sampling rates to be better (at the cost of slower performance), but we haven't tested this rigorously.  
\end{itemize}

Other hyperparameters are less obvious. \textit{Number of clusters} in the approximate matching step is likely good in moderation. While increasing it will give us more landmark points for alignment, it can theoretically lead to instabilities because the assignment of very granular centroids can be noisy. Increasing \textit{number of neighbors} is often beneficial, but it eventually leads to diminishing returns, and might even deteriorate performance when neighborhoods become very large. At moderate scales, it helps to reduce noise, as averaging more neighbors leads to a smoother estimate of the true aligned position. Note that this is more relevant here than in word embeddings, where essentially every point has (more or less) one ground-truth translation and it makes sense to rely on a single neighbor. Increasing the number of neighbors to very large numbers might lead to over-smoothing, which may harm performance.  
Finally, increasing the exponential smoothing hyperparameter $\alpha$ leads to higher cosine scores, but they don't necessarily translate to better rank and top-1 scores. $\alpha$ is implicitly responsible for the distance of the matrix from being orthogonal. Additionally, increasing $\alpha$ makes convergence slightly slower. 
 
In summary, unlike many deep learning setups, GANs in particular, the range of hyperparameters that lead to strong performance is generously wide. Each hyperparameter is interpretable, and one can plausibly guess in which direction to modify the parameter to improve effectiveness, if needed.

\end{document}